\PassOptionsToPackage{dvipsnames,svgnames,table}{xcolor}
\documentclass[anon]{l4dc2027}
\usepackage{times}
\usepackage{xurl}
\hypersetup{
    colorlinks,
    linkcolor={red!50!black},
    citecolor={blue!50!black},
    urlcolor={blue!80!black}
}
\usepackage{bm}

\usepackage{enumitem}
\usepackage{array}
\usepackage{booktabs}
\usepackage{dcolumn}
\newcolumntype{d}[1]{D..{#1}}
\usepackage{soul}
\usepackage{makecell}
\hypersetup{
  colorlinks,
  citecolor=teal,
  linkcolor=teal,
  urlcolor=purple,
}
\usepackage[font=scriptsize,labelfont=bf]{caption}
\LinesNumbered
\SetAlgoVlined
\usepackage[normalem]{ulem}
\newcommand{\reals}{\mathbb{R}}

\newcommand{\R}{\reals}
\newcommand{\Rnonneg}{\reals_{\geq 0}}
\newcommand{\Rplus}{\reals_{>0}}

\newcommand{\naturals}{\mathbb{N}}
\renewcommand{\S}{\mathbb{S}}
\newcommand{\indicator}{\mathbb{I}}

\newcommand{\E}{\mathbb{E}}

\newcommand{\Bcal}{\mathcal{B}}

\newcommand{\Dcal}{\mathcal{D}}

\newcommand{\Fcal}{\mathcal{F}}

\newcommand{\Ncal}{\mathcal{N}}

\newcommand{\Pcal}{\mathcal{P}}

\newcommand{\Scal}{\mathcal{S}}
\newcommand{\Tcal}{\mathcal{T}}
\newcommand{\Ucal}{\mathcal{U}}

\newcommand{\Xcal}{\mathcal{X}}

\newcommand{\Zcal}{\mathcal{Z}}

\DeclareMathOperator*{\argmax}{arg\,max}

\definecolor{yellow}{cmyk}{0.0,0.10,0.95,0.0}
\definecolor{pred}{cmyk}{0,0.8,0.70,0.0}
\definecolor{bluedefined}{cmyk}{0.46,0.10,0,0.0}
\usepackage{mdframed}
\newmdenv[
  linewidth=0.8pt, topline=false, bottomline=false,
  rightline=true, leftline=true, innertopmargin=4pt,
  innerbottommargin=4pt, innerleftmargin=8pt, innerrightmargin=8pt
]{problembar}
\usepackage[most]{tcolorbox}
\definecolor{problemblue}{RGB}{239,247,252}
\definecolor{problemline}{RGB}{210,220,235}
\definecolor{problemgray}{RGB}{247,247,247}
\definecolor{problemgrayline}{RGB}{210,210,210}
\definecolor{problemred}{RGB}{255,245,245}
\definecolor{problemredline}{RGB}{220,160,160}
\definecolor{proofblue}{RGB}{247,251,254}
\definecolor{proofline}{RGB}{225,232,242}
\newtcolorbox{highlight}{
  enhanced, breakable, colback=problemgray, colframe=problemgray,
  boxrule=0pt, frame hidden,
  borderline west={3pt}{0pt}{problemgrayline}, arc=0pt, outer arc=0pt,
  boxsep=0pt, left=7pt, right=6pt, top=6pt, bottom=6pt
}
\newtcolorbox{banner}{
  enhanced, breakable, colback=white, colframe=white,
  boxrule=0pt, frame hidden,
  borderline west={3pt}{0pt}{problemgrayline}, arc=0pt, outer arc=0pt,
  boxsep=0pt, left=7pt, right=0pt, top=0pt, bottom=0pt
}
\newtcolorbox{highlight_red}{
  enhanced, breakable, colback=problemred, colframe=problemred,
  boxrule=0pt, frame hidden,
  borderline west={3pt}{0pt}{problemredline}, arc=0pt, outer arc=0pt,
  boxsep=0pt, left=7pt, right=6pt, top=6pt, bottom=6pt
}
\newtcolorbox{highlight_blue}{
  enhanced, breakable, colback=problemblue, colframe=problemblue,
  boxrule=0pt, frame hidden,
  borderline west={3pt}{0pt}{problemline}, arc=0pt, outer arc=0pt,
  boxsep=0pt, left=7pt, right=6pt, top=6pt, bottom=6pt
}
\newtcolorbox{highlight_blue_proof}{
  enhanced, breakable, colback=proofblue, colframe=proofblue,
  boxrule=0pt, frame hidden,
  borderline west={3pt}{0pt}{proofline}, arc=0pt, outer arc=0pt,
  boxsep=0pt, left=7pt, right=6pt, top=6pt, bottom=6pt
}

\newif\ifproofreading
\proofreadingtrue
\DeclareRobustCommand{\beforefix}[1]{\ifproofreading\ifmmode\else\leavevmode\fi{\color{red}\hypersetup{allcolors=red}#1}\fi}
\DeclareRobustCommand{\fixed}[1]{\ifproofreading\ifmmode\else\leavevmode\fi{\color{blue}\hypersetup{allcolors=blue}#1}\else#1\fi}
\newtheorem{problem}{Problem}

\title[Disturbance-Aware Information Planning]{Disturbance-Aware Informative Planning in Spatiotemporal Environments}
\author{\Name{Anonymous Authors}}
\author{%
 \Name{Kaleb Ben Naveed}\Email{kbnaveed@umich.edu}\\
 \addr Department of Robotics, University of Michigan, Ann Arbor
 \AND
  \Name{Haejoon Lee}\Email{haejoonl@umich.edu}\\
 \addr Department of Robotics, University of Michigan, Ann Arbor
 \AND
 \Name{Anouck Girard}\Email{girarda3@erau.edu}\\
 \addr Embry-Riddle Aeronautical University
 \AND
 \Name{Dimitra Panagou}\Email{dpanagou@umich.edu}\\
 \addr Department of Robotics, Department of Aerospace Engineering, University of Michigan, Ann Arbor
}

\begin{document}
\maketitle

\begin{abstract}
Autonomous robots are widely used for persistent monitoring of spatiotemporal environments, where quantities of interest (QoI) change across space and time and require repeated measurements. In practice, robot motion can be affected by external disturbances that vary across the environment. When such external disturbances are not considered in trajectory optimization, it can lead to inaccurate predictions of robot motion and, consequently, the information acquired about the QoI during execution. However, estimating disturbances throughout the environment can be inefficient, as disturbances in regions with little overlap with monitoring trajectories have limited impact on monitoring performance. We therefore propose a two-stage informative planning framework for unknown, spatially varying but time-invariant disturbances, combining informed disturbance sampling with disturbance-aware trajectory optimization. First, we identify where QoI measurements are needed and generate alternative paths to these locations using the environment geometry. We then prioritize disturbance measurements based on their expected reduction of disturbance uncertainty along these paths. Second, rather than optimizing information acquisition based on nominal robot motion, we incorporate the learned disturbance belief into trajectory optimization to account for its effects on robot motion and information acquisition. Across four simulation scenarios, the proposed method reduces disturbance-learning time by $25.0$--$52.3\%$ compared with environment-wide estimation and identifies paths that appear costly under nominal dynamics but become preferable when disturbances are considered. \href{https://kalebbennaveed.github.io/dist-aware-ipp/}{[Project Page]}\footnote{Project Page: \url{https://kalebbennaveed.github.io/dist-aware-ipp/}}.

\end{abstract}

\section{Introduction}
\label{sec:introduction}
Autonomous robots are widely used for environmental monitoring, collecting measurements of quantities of interest such as temperature, water quality, and terrain characteristics~\citep{hitz2017adaptive,binney2013waypoints,popovic2020terrain}. In spatiotemporal environments, these quantities of interest evolve over time, causing information decay. Informative planning must therefore account for both information acquisition and decay when optimizing robot trajectories. 
A separate challenge is uncertainty about external disturbances, which can affect robot motion and the locations where measurements are collected. Consequently, trajectories optimized using nominal dynamics may acquire less information than predicted during execution.

Addressing these challenges requires accounting for uncertainty about the quantities of interest being monitored and uncertainty about the external disturbances affecting robot motion. Informative planning methods quantify uncertainty in the quantities of interest using measures such as posterior variance, entropy, mutual information, and clarity~\citep{krause2008sensors,singh2007paths,agrawal2023clarity}. These measures guide the selection of sensing locations and trajectories through sampling-based planning under travel budgets~\citep{singh2007paths,hollinger2014sampling} or optimization of expected information gain~\citep{hitz2017adaptive,popovic2020terrain,marchant2014bayesian}. In spatiotemporal environments, informative planning must account for evolving quantities of interest. Existing approaches use spatiotemporal Gaussian processes~\citep{binney2013waypoints}, update sparse spatial GP models online~\citep{ma2017sparse}, or explicitly model information acquisition and decay in ergodic exploration and trajectory optimization~\citep{naveed2024eclares,naveed2026stein}. However, they do not jointly address disturbance learning and its effects on robot motion during informative planning.

Beyond uncertainty in the quantities of interest, other approaches account for the effects of disturbances on robot motion. \citet{popovic2020localization} consider localization and mapping uncertainty in informative planning, while \citet{cho2018wind} incorporate known wind through direction-dependent travel costs. For unknown disturbances, \citet{sun2021safe} use Gaussian processes to learn state-dependent disturbances for safe exploration. Other approaches combine online wind estimation with wind-aware routing~\citep{moon2025thesis} or use learned wind maps in quadrotor planning and control to reduce attitude effort~\citep{torok2026wind}. While these methods address disturbance estimation or its effects on robot motion, they do not jointly consider monitoring-driven disturbance learning and the incorporation of learned disturbance beliefs into informative trajectory optimization.

Selecting disturbance measurements to improve subsequent monitoring requires considering how reducing disturbance uncertainty affects future planning decisions. This is closely related to dual control, which balances immediate performance and learning~\citep{arcari2020dual,li2025dual}, and task-optimal exploration, which prioritizes learning according to its effect on task performance~\citep{wagenmaker2021task,wagenmaker2023nonlinear}. However, directly evaluating this benefit is computationally challenging, as it requires considering possible measurement outcomes, their effects on the disturbance belief, and the resulting changes in subsequent informative trajectories.

To address these challenges, we propose a two-stage disturbance-aware informative planning framework. \textcolor{magenta}{(i)}
The first stage predicts where QoI measurements are needed as information decays and generates alternative paths to these locations using environment geometry. We then prioritize disturbance measurements by their expected reduction of disturbance uncertainty along these paths, weighted by monitoring relevance and accounting for travel cost. \textcolor{magenta}{(ii)}
The second stage incorporates the learned disturbance belief into Model-Based Diffusion (MBD) trajectory optimization, so that the trajectory cost reflects information acquisition and decay along predicted robot motion under disturbances rather than nominal motion. We compare against nominal planning, disturbance-aware monitoring without dedicated learning, and environment-wide disturbance estimation, evaluating monitoring performance over both stages.

\section{Preliminaries}
\label{sec:preliminaries}
\subsection{Notation}
Let $\naturals=\{0,1,\ldots\}$ and $\naturals_{>0}=\{1,2,\ldots\}$; $\R$, $\Rnonneg$, and $\Rplus$ denote the real, nonnegative, and positive real numbers. For $k\in\naturals_{>0}$, $I_k$, $\S_+^k$, and $\S_{++}^k$ denote the identity matrix and the sets of symmetric positive-semidefinite and positive-definite matrices in $\R^{k\times k}$, respectively. We write $\Ncal(\mu,\Sigma)$ for a Gaussian distribution with mean $\mu$ and covariance $\Sigma$, $\operatorname{tr}(\cdot)$ for the trace, $[a]_+=\max\{a,0\}$ for $a\in\R$, and $\indicator_A(z)$ for the indicator of $z\in A$.

\subsection{Dynamics Model}
\label{sec:dynamics}
Consider the continuous-time robot dynamics
\begin{equation}
 \dot x=f\bigl(x,u,\bar d(\Psi(x))\bigr), \qquad t\geq t_0,
 \label{eq:dynamics}
\end{equation}
where $x\in\Xcal\subset\R^n$ and $u\in\Ucal\subset\R^m$ denote the robot state and control input. The robot's position is given by $\Psi:\Xcal\rightarrow\Pcal$, where $\Pcal\subset\R^{n_r}$ is the workspace and $n_r\in\naturals_{>0}$. The unknown disturbance field is $\bar d:\Pcal\rightarrow\R^{n_d}$, with $n_d\in\naturals_{>0}$, and $f:\Xcal\times\Ucal\times\R^{n_d}\rightarrow\R^n$ is the known dynamics model. The field $\bar d$ can vary with position, but its value at any fixed $r\in\Pcal$ does not change with time. Let $\Scal\subset\Xcal$ denote the prescribed set of admissible states, including obstacle avoidance.

\begin{definition}[Trajectory]
Let $\Tcal=[t_i,t_f]\subset\R$, where $t_i<t_f$.
A trajectory starting from $(t_i,x_i)\in\R\times\Xcal$
is a pair of functions
$\xi=(\xi_x:\Tcal\rightarrow\Xcal,
\xi_u:\Tcal\rightarrow\Ucal)$ satisfying
$
\dot\xi_x(t)=f\bigl(\xi_x(t),\xi_u(t),
\bar d(\Psi(\xi_x(t)))\bigr),
\quad \forall t\in(t_i,t_f),
$
with $\xi_x(t_i)=x_i$.
\end{definition}

\subsection{Quantity of Interest Model and Clarity}
\label{sec:clarity}
Let $N_p\in\naturals_{>0}$ be the number of free-space monitoring cells and $\Pcal_{\mathrm{cells}}=\{1,\ldots,N_p\}$ their index set. Cell $p\in\Pcal_{\mathrm{cells}}$ has center $r_p\in\Pcal$ and a scalar quantity of interest $m_p(t)\in\R$. The cell quantities have independent Gaussian initial beliefs. Their process and observation models are
\begin{equation}
 \begin{aligned}
 dm_p(t)&=\sqrt{Q_p}\,dW_p(t),\\
 dy_p(t)&=C_p(x)m_p(t)\,dt+\sqrt{R}\,dV_p(t),
 \end{aligned}
 \label{eq:environment_model}
\end{equation}
where $Q_p\in\Rnonneg$ and $R\in\Rplus$ are process- and measurement-noise intensities. The scalar processes $W_p,V_p$ are mutually independent standard Wiener processes, and $y_p(t)\in\R$ is the integrated observation. The function $C_p:\Xcal\rightarrow\R$ specifies the measurement sensitivity to cell $p$ and is zero when that cell is not observed. Different values of $Q_p$ allow the quantities at different cells to evolve with different process uncertainty.

Estimating each $m_p$ with a continuous-time Kalman filter gives posterior mean $\hat m_p(t)\in\R$ and variance $P_p(t)\in\Rnonneg$, with $\dot P_p=Q_p-C_p(x)^2P_p^2/R$. Clarity~\citep{agrawal2023clarity} is a normalized representation of differential entropy; for this scalar Gaussian belief, $q_p=(1+P_p)^{-1}\in(0,1]$. Substitution of the variance equation yields
\begin{equation}
 \dot q_p=g_p(x,q_p):=\frac{C_p(x)^2}{R}(1-q_p)^2-Q_pq_p^2,
 \qquad g_p:\Xcal\times[0,1]\rightarrow\R.
 \label{eq:clarity_dynamics}
\end{equation}
The first term represents information acquisition and the second represents information decay. Let $\bar q_p\in[0,1]$ be the target clarity in cell $p$. With $q(t)=[q_p(t)]_{p=1}^{N_p}\in(0,1]^{N_p}$, the cellwise deficits and mean-deficit function $q_d:(0,1]^{N_p}\rightarrow[0,1]$ are
\begin{equation}
 \delta_p(t)=[\bar q_p-q_p(t)]_+\in[0,1],\qquad
 q_d(q(t))=\frac{1}{N_p}\sum_{p=1}^{N_p}\delta_p(t)\in[0,1].
 \label{eq:clarity_deficit}
\end{equation}
Thus, $\delta_p$ measures the difference between target and current clarity. 

\subsection{Disturbance Belief}
\label{sec:disturbance}
To estimate the unknown disturbance field $\bar d$, the robot collects noisy measurements along its trajectory. Let $t_j^d\geq t_0$ denote the time of the $j$-th disturbance measurement, where $j\in\naturals_{>0}$ indexes successive measurements. At position $r_j=\Psi(x(t_j^d))\in\Pcal$, the measurement $y_j^d\in\R^{n_d}$ is given by
\begin{equation}
 y_j^d=\bar d(r_j)+\nu_j^d,
 \qquad \nu_j^d\sim\Ncal(0,R_d),
 \label{eq:disturbance_observation}
\end{equation}
where $\nu_j^d\in\R^{n_d}$ is the measurement noise with covariance $R_d\in\S_{++}^{n_d}$. The measurement noises are independent across $j$ and independent of the noises in~\eqref{eq:environment_model}. The dataset $\Dcal_t^d=\{(r_j,y_j^d):t_0\leq t_j^d\leq t\}$ contains all disturbance measurements collected up to time $t$.

To represent uncertainty about the unknown field $\bar d$, we consider a random field $d:\Pcal\rightarrow\R^{n_d}$. 
Let $\Bcal_t^d$ denote its posterior belief conditioned on $\Dcal_t^d$, and let $\mathcal B_d$ be the space of such disturbance beliefs.
 For any $r,r'\in\Pcal$, the posterior mean and cross-covariance are
\begin{equation}
 \begin{aligned}
 \mu_d(r,t)
 &=\E_{\Bcal_t^d}[d(r)]\in\R^{n_d},\\
 \Sigma_d(r,r',t)
 &=\operatorname{Cov}_{\Bcal_t^d}[d(r),d(r')]
 \in\R^{n_d\times n_d}.
 \end{aligned}
 \label{eq:disturbance_moments}
\end{equation}
The marginal covariance is $P_d(r,t)=\Sigma_d(r,r,t)\in\S_+^{n_d}$. Although the physical disturbance field $\bar d$ is time-invariant, its posterior belief $\Bcal_t^d$ is updated as new measurements are collected.

\subsection{Probabilistic Online Attentive Mapping (POAM) for Estimating Disturbance}
\label{sec:poam}
We estimate the unknown disturbance field using Probabilistic Online Attentive Mapping (POAM)~\citep{chen2024poam}, an online sparse Gaussian-process (GP) method. Its attentive kernel combines base kernels of different lengthscales using location-dependent weights, allowing spatially varying correlations. 
We approximate $\Bcal_t^d$ by independent sparse GP posteriors for the disturbance components and assume diagonal observation-noise covariance $R_d$.

For component $d_a:\Pcal\rightarrow\R$, $a\in\{1,\ldots,n_d\}$, we use a zero-mean GP prior with kernel
 $k_{a,t}:\Pcal\times\Pcal\rightarrow\R$, whose parameters are updated as new measurements become available. Let $M_a\in\naturals_{>0}$ and $Z_{a,t}=(z_{a,1,t},\ldots,z_{a,M_a,t})\in\Pcal^{M_a}$ denote the number and locations of inducing points. The corresponding inducing values $\zeta_{a,t}=[d_a(z_{a,1,t}),\ldots,d_a(z_{a,M_a,t})]^\top\in\R^{M_a}$ have variational posterior $\Ncal(m_{a,t}^\zeta,S_{a,t})$, where $m_{a,t}^\zeta\in\R^{M_a}$ and $S_{a,t}\in\S_{++}^{M_a}$. Define the nonsingular kernel matrix $K_{a,t}=k_{a,t}(Z_{a,t},Z_{a,t})\in\S_{++}^{M_a}$ and $h_{a,t}(r)=K_{a,t}^{-1}k_{a,t}(Z_{a,t},r)\in\R^{M_a}$, where $k_{a,t}(Z_{a,t},r)\in\R^{M_a}$ contains the kernel values between the inducing points and $r$. For any $r,r'\in\Pcal$, the predictive mean and covariance are
\begin{equation}
 \begin{aligned}
 \mu_{d,a}(r,t)
 &=h_{a,t}(r)^\top m_{a,t}^\zeta,\\
 \Sigma_{d,a}(r,r',t)
 &=k_{a,t}(r,r')
 -k_{a,t}(r,Z_{a,t})h_{a,t}(r')
 +h_{a,t}(r)^\top S_{a,t}h_{a,t}(r').
 \end{aligned}
 \label{eq:poam_prediction}
\end{equation}
The vector mean $\mu_d(r,t)\in\R^{n_d}$ consists of the component means $\mu_{d,a}(r,t)\in\R$, while the cross-covariance $\Sigma_d(r,r',t)\in\R^{n_d\times n_d}$ has diagonal entries $\Sigma_{d,a}(r,r',t)\in\R$ and zero cross-component entries. POAM updates inducing locations using 
pivoted Cholesky decomposition (PCD)%
, incorporates new measurements through analytic variational updates, and adapts kernel hyperparameters through mini-batch optimization. For further details, see~\citet{chen2024poam}.

\subsection{Model-Based Diffusion}
\label{sec:mbd}

Model-Based Diffusion (MBD)~\citep{pan2024mbd} is a zeroth-order, sampling-based optimization method that iteratively refines candidate solutions through score-based reverse diffusion using samples, without requiring objective gradients.

Let $\theta\in\Theta\subset\R^{n_\theta}$, with $n_\theta\in\naturals_{>0}$, denote the decision vector and $J:\Theta\rightarrow\R$ its cost. MBD defines $p_0(\theta)\propto\exp(-J(\theta)/\lambda)$ on $\Theta$ and $p_0=0$ outside $\Theta$, with temperature $\lambda\in\Rplus$. For $N_D\in\naturals_{>0}$ diffusion steps, indexed by $\nu=1,\ldots,N_D$, define $\alpha_\nu\in(0,1)$ and $\bar\alpha_\nu=\prod_{k=1}^{\nu}\alpha_k$. At each step, $Y^{(\nu)}\in\R^{n_\theta}$ denotes a noisy version of the decision vector $\theta$, generated by the forward diffusion process:
\begin{equation}
 Y^{(\nu)}\mid\theta\sim
 \Ncal\left(\sqrt{\bar\alpha_\nu}\theta,
 (1-\bar\alpha_\nu)I_{n_\theta}\right).
 \label{eq:mbd_forward}
\end{equation}
During reverse diffusion, $K_\theta\in\naturals_{>0}$ proposals are sampled and weighted by their costs:
\begin{equation}
\theta^{(b)}\sim\Ncal\left(
\frac{Y^{(\nu)}}{\sqrt{\bar\alpha_\nu}},
\frac{1-\bar\alpha_\nu}{\bar\alpha_\nu}I_{n_\theta}
\right),\quad b=1,\ldots,K_\theta,
\qquad
\widehat\theta_\nu=
\frac{\sum_{b=1}^{K_\theta}
\theta^{(b)}p_0(\theta^{(b)})}
{\sum_{b=1}^{K_\theta}p_0(\theta^{(b)})}.
\label{eq:mbd_proposals}
\end{equation}
For a positive denominator, $\widehat\theta_\nu\in\R^{n_\theta}$ estimates the clean decision conditioned on $Y^{(\nu)}$. Let $p_\nu$ be the marginal density of $Y^{(\nu)}$ for $\theta\sim p_0$. Its score $\nabla\log p_\nu(Y^{(\nu)})$ is approximated by $\widehat s_\nu\in\R^{n_\theta}$ in the reverse update:
\begin{equation}
 \widehat s_\nu=\frac{-Y^{(\nu)}+\sqrt{\bar\alpha_\nu}\widehat\theta_\nu}{1-\bar\alpha_\nu},
 \qquad
 Y^{(\nu-1)}=\frac{Y^{(\nu)}+(1-\bar\alpha_\nu)\widehat s_\nu}{\sqrt{\alpha_\nu}}.
 \label{eq:mbd_reverse}
\end{equation}

Repeating this update for $\nu=N_D,\ldots,1$ progressively refines the candidates toward lower-cost solutions. Further details are provided by~\citet{pan2024mbd}.

\section{Problem Statement}
\label{sec:problem}
Consider a robot governed by~\eqref{eq:dynamics} monitoring~\eqref{eq:environment_model}, with initial state $x_0\in\Scal$, clarity $q_0\in(0,1]^{N_p}$, 
and disturbance belief $\Bcal_0^d=\Bcal_{t_0}^d$ at $t_0$. A feedback policy $\pi:\Xcal\times(0,1]^{N_p}\times\mathcal B_d\rightarrow\Ucal$ selects $u=\pi(x,q,\Bcal_t^d)$ without access to the true field or future measurements.
 We seek to minimize the expected long-run mean clarity deficit subject to state and input constraints:
\begin{subequations}
\label{eq:persistent_problem}
\begin{align}
 \min_{\pi}\quad
 &\limsup_{T\rightarrow\infty}\frac{1}{T}\E\!\left[
       \int_{t_0}^{t_0+T}q_d(q(\tau))\,d\tau\right],
 \label{eq:persistent_objective}\\
 \mathrm{s.t.}\quad
 &\dot x=f\bigl(x,u,\bar d(\Psi(x))\bigr),\\
 &\dot q_p=g_p(x,q_p),
       \quad p\in\Pcal_{\mathrm{cells}},\\
 &u=\pi(x,q,\Bcal_t^d)\in\Ucal,\qquad x\in\Scal,
       \quad t\geq t_0,\\
 &x(t_0)=x_0,\qquad q(t_0)=q_0.
\end{align}
\end{subequations}
Here $T\in\Rplus$ is the averaging duration, not a terminal mission time. The expectation includes the initial disturbance uncertainty and observation noise; constraints are required almost surely for all $t\geq t_0$. The belief $\Bcal_t^d$ is updated from observations~\eqref{eq:disturbance_observation}.

Directly solving~\eqref{eq:persistent_problem} is computationally intractable, as it requires optimizing robot trajectories while accounting for future disturbance measurements and belief updates over an infinite horizon. We therefore consider the following problem and propose a safety-aware approximate solution to it.

\textbf{\begin{figure}[!ht]
\centering
\includegraphics[width=\linewidth,height=0.78\textheight,keepaspectratio]{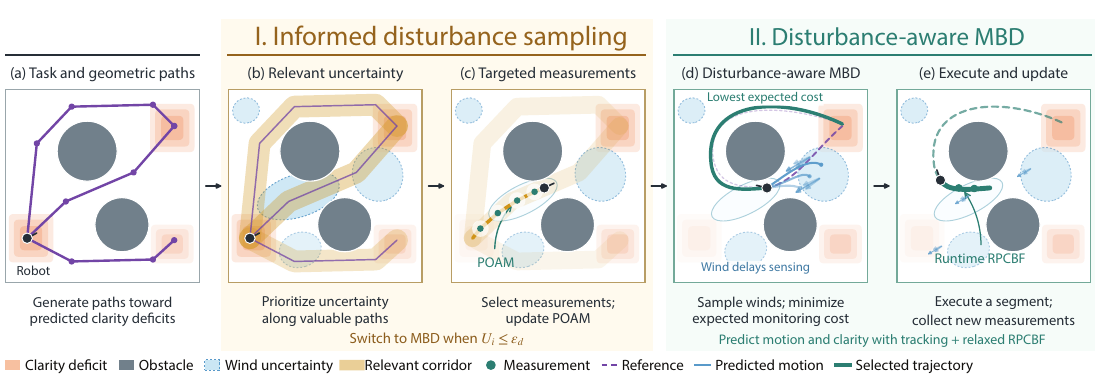}
\caption{Overview of monitoring-informed disturbance estimation and disturbance-aware informative planning.}
\vspace{-5mm}
\label{fig:proposed_method}
\end{figure}}

\begin{problem}[Disturbance-aware informative planning]
\label{prob:disturbance_aware}
Given the robot and monitoring models, environment geometry, and local disturbance measurements~\eqref{eq:disturbance_observation}, develop an online method to approximate the constrained monitoring problem~\eqref{eq:persistent_problem} by (i) learning the unknown disturbance field in regions relevant to the monitoring task and (ii) incorporating the learned disturbance belief into informative trajectory optimization while accounting for state and input constraints.
\end{problem}

\section{Methodology}
\label{sec:method}
\subsection{Overview}
\label{sec:overview}

We develop the two-stage online approach to Problem~\ref{prob:disturbance_aware} illustrated in Fig.~\ref{fig:proposed_method}.
 The first stage collects disturbance measurements in regions relevant to the monitoring task. Once the uncertainty in these regions is sufficiently reduced, the second stage uses the learned disturbance belief to optimize informative trajectories. Measurements of both the quantities of interest and the disturbance continue throughout both stages.

Let $t_i$, $i\in\naturals$, denote successive planning times, with $t_{i+1}>t_i$ and $t_i\rightarrow\infty$. We define $x_i=x(t_i)\in\Xcal$, $q_i=[q_p(t_i)]_{p=1}^{N_p}\in(0,1]^{N_p}$, and $\Bcal_i^d=\Bcal_{t_i}^d$ as the current robot state, clarity vector, and disturbance belief, respectively. The clarity targets are $\bar q=[\bar q_p]_{p=1}^{N_p}\in[0,1]^{N_p}$. Trajectories are planned over a horizon $T_N\in\Rplus$ and executed in a receding-horizon manner. The planning index $i$ continues across both stages.

\subsection{Geometric Path Initialization}
\label{sec:geometric_init}
We first generate geometric paths toward informative sensing locations using the current clarity map and known obstacle geometry. These paths identify regions where disturbance estimation can benefit the monitoring task and provide distinct initial trajectories for subsequent optimization.

Since the quantities of interest evolve over time, selecting sensing locations based only on the current clarity may overlook regions with significant future clarity deficits. We therefore predict clarity over $[t_i,t_i+T_N]$, assuming no additional measurements are collected. Setting $C_p=0$ in~\eqref{eq:clarity_dynamics}, the predicted clarity $q_{i,p}^{\mathrm{ns}}(\tau)$ of cell $p\in\Pcal_{\mathrm{cells}}$ at time $t_i+\tau$, for $\tau\in[0,T_N]$, and its corresponding deficit are
\begin{equation}
 q_{i,p}^{\mathrm{ns}}(\tau)
 =\frac{q_p(t_i)}{1+Q_pq_p(t_i)\tau},
 \qquad
 \delta_{i,p}^{\mathrm{ns}}(\tau)
 =[\bar q_p-q_{i,p}^{\mathrm{ns}}(\tau)]_+.
 \label{eq:deficit_preview}
\end{equation}
We evaluate these deficits at $N_\tau\in\naturals_{>0}$ selected times $\tau_h\in[0,T_N]$, $h=1,\ldots,N_\tau$, forming $D_i^{\mathrm{ns}}=[\delta_{i,p}^{\mathrm{ns}}(\tau_h)]_{p,h}\in[0,1]^{N_p\times N_\tau}$. This accounts for how clarity deficits develop throughout the prediction horizon.

Using these predictions and known obstacle geometry, we generate a fixed number $N_\gamma\in\naturals_{>0}$ of alternative collision-free paths. Let $\Pcal_{\mathrm{free}}\subset\Pcal$ denote the collision-free workspace and $\Gamma_i=\{\gamma_i^\ell\}_{\ell=1}^{N_\gamma}$ the generated paths, where $\ell=1,\ldots,N_\gamma$ indexes individual paths. Each path $\gamma_i^\ell:[0,1]\rightarrow\Pcal_{\mathrm{free}}$ starts at $\Psi(x_i)$ and reaches a candidate sensing location, with $s\in[0,1]$ denoting normalized arc length. Alternative paths through different passages are retained to provide geometrically distinct routes.

To prioritize paths according to their monitoring relevance, we evaluate the predicted clarity deficits of cells observable along each path. Let $x_i^\ell(s)\in\Xcal$ be a sensing configuration at $\gamma_i^\ell(s)$, satisfying $\Psi(x_i^\ell(s))=\gamma_i^\ell(s)$. The score of the $\ell$-th path is defined as
\begin{equation}
 a_i^\ell=\frac{1}{N_\tau}\sum_{h=1}^{N_\tau}
 \int_0^1\sum_{p=1}^{N_p}
 C_p(x_i^\ell(s))^2
 \delta_{i,p}^{\mathrm{ns}}(\tau_h)\,ds.
 \label{eq:path_score}
\end{equation}
The integral can be approximated numerically by evaluating the integrand at discrete points along each path. A larger score indicates that the path passes through locations from which cells with greater predicted clarity deficits can be observed. These scores form $a_i=[a_i^\ell]_{\ell=1}^{N_\gamma}\in\Rnonneg^{N_\gamma}$ and are subsequently used to prioritize disturbance estimation along the geometric paths. The generated paths are also used to initialize the subsequent MBD trajectory optimization.

\subsection{Disturbance Estimation}
\label{sec:estimation}
We use the geometric paths and their monitoring relevance scores from Sec.~\ref{sec:geometric_init} to determine where disturbance measurements should be prioritized.

\paragraph{Monitoring-informed spatial weighting.}
Disturbances along geometric paths can affect the robot's motion and its ability to collect measurements of the quantities of interest. We therefore construct spatial weights that assign greater importance to regions near paths with higher monitoring relevance. Let $\kappa_\rho(r,\gamma)\in\Rnonneg$ be a prescribed spatial weighting function that assigns greater weight to locations near the geometric path $\gamma$, where $r\in\Pcal_{\mathrm{free}}$ and $\rho\in\Rplus$ is a tunable parameter. 
For each cell $p\in\Pcal_{\mathrm{cells}}$, define an unnormalized weight and, when their sum is positive, its normalized value by
\begin{equation}
 \widetilde\omega_{i,p}
 =\sum_{\ell=1}^{N_\gamma}
 a_i^\ell\kappa_\rho(r_p,\gamma_i^\ell),
 \qquad
 \omega_{i,p}=\frac{\widetilde\omega_{i,p}}{W_i},
 \label{eq:learning_weights}
\end{equation}
where $W_i=\sum_{p=1}^{N_p}\widetilde\omega_{i,p}$ is the normalization factor. For $W_i>0$, the weights $\omega_i=[\omega_{i,p}]_{p=1}^{N_p}\in\Rnonneg^{N_p}$ sum to one. 

\paragraph{Disturbance measurement selection.}
Using the spatial weights, we quantify the disturbance uncertainty in regions relevant to the monitoring task as
\begin{equation}
 U_i=\left(\sum_{p=1}^{N_p}\omega_{i,p}\operatorname{tr}P_d(r_p,t_i)\right)^{1/2}.
 \label{eq:weighted_uncertainty}
\end{equation}
We aim to select measurements that reduce this uncertainty. Since the disturbance belief captures spatial correlations, a measurement at one location can reduce uncertainty at other locations. For a candidate measurement location $z\in\Pcal_{\mathrm{free}}$, let $G_i(z)\in\Rnonneg$ denote the expected reduction in $U_i^2$ from a disturbance measurement at $z$. Using Gaussian conditioning, this reduction is given by
\begin{equation}
G_i(z)=\sum_{p=1}^{N_p}\omega_{i,p}
\operatorname{tr}\left[
\Sigma_d(r_p,z,t_i)
(P_d(z,t_i)+R_d)^{-1}
\Sigma_d(z,r_p,t_i)
\right].
\label{eq:measurement_gain}
\end{equation}

This one-measurement reduction is evaluated with the current spatial weights, GP kernel parameters, and inducing representation held fixed; subsequent POAM retraining is not predicted.
 To account for travel cost, we consider a finite set of candidate measurement locations $\Zcal_i\subset\Pcal_{\mathrm{free}}$. Let $L_i:\Zcal_i\rightarrow\Rnonneg$ give the collision-free travel distance from $\Psi(x_i)$ to each candidate, and let $L_0\in\Rplus$ control the travel-distance penalty. When candidates remain after route checks, we select
\begin{equation}
 z_i^\star\in\argmax_{z\in\Zcal_i}
 \frac{G_i(z)}{1+L_i(z)/L_0}.
 \label{eq:measurement_selection}
\end{equation}
We exclude candidate routes that intersect obstacles.
\paragraph{Stopping criterion for disturbance estimation.}
At each planning iteration, $U_i$ is updated using the current clarity and disturbance belief. Dedicated disturbance estimation terminates when $U_i\leq\varepsilon_d$, where $\varepsilon_d\in\Rplus$ is a prescribed uncertainty tolerance. Throughout this stage, clarity continues to evolve, and the monitoring cost incurred during disturbance estimation is included in the evaluation.

\subsection{Disturbance-Aware Model-Based Diffusion}
\label{sec:disturbance_mbd}
In the second stage, we incorporate the learned disturbance belief into Model-Based Diffusion (MBD) to optimize informative trajectories. At each planning time $t_i$, geometric paths from Sec.~\ref{sec:geometric_init} initialize candidate trajectories, which are refined through MBD. Each candidate is evaluated by predicting the robot's motion under sampled disturbance fields and the resulting information acquisition, allowing the optimizer to account for disturbance effects directly.

\paragraph{Disturbance-aware trajectory prediction.}

Let $\theta\in\Theta\subset\R^{n_\theta}$ parameterize candidate state and input curves $\bar\xi(\cdot;\theta)=(\bar\xi_x,\bar\xi_u)$ over $[t_i,t_i+T_N]$.
 To evaluate each candidate, we draw $N_d\in\naturals_{>0}$ spatial disturbance fields $d_i^{(s)}:\Pcal\rightarrow\R^{n_d}$ from $\Bcal_i^d$, indexed by $s=1,\ldots,N_d$. Each sampled field remains fixed throughout its rollout, preserving spatial correlations.

Let $\mathcal T_i=\{\bar\xi(\cdot;\theta):\theta\in\Theta\}$ denote the set of candidate trajectories.
 For each sample, the tracking controller $\pi_{\mathrm{tr}}:[t_i,t_i+T_N]\times\Xcal\times\mathcal T_i\times\mathcal B_d\rightarrow\R^m$ computes a nominal control input to follow $\bar\xi(\cdot;\theta)$ using the current predicted state and disturbance belief.
 This input is modified by an approximate safety filter $\Fcal^{\mathrm{pred}}:\Xcal\times\R^m\times\mathcal B_d\rightarrow\Ucal$, based on the approach described in the following paragraph. The resulting predicted state $\xi_x^{\theta,s}(t)\in\R^n$ and input $\xi_u^{\theta,s}(t)\in\R^m$ satisfy
\begin{equation}
 \begin{aligned}
 u_{\mathrm{nom}}^{\theta,s}(t)
 &=\pi_{\mathrm{tr}}(t,\xi_x^{\theta,s}(t),
 \bar\xi(\cdot;\theta),\Bcal_i^d),\\
 \xi_u^{\theta,s}(t)
 &=\Fcal^{\mathrm{pred}}(\xi_x^{\theta,s}(t),
 u_{\mathrm{nom}}^{\theta,s}(t),\Bcal_i^d),\\
 \dot\xi_x^{\theta,s}(t)
 &=f\bigl(\xi_x^{\theta,s}(t),\xi_u^{\theta,s}(t),
 d_i^{(s)}(\Psi(\xi_x^{\theta,s}(t)))\bigr),
 \end{aligned}
 \label{eq:controlled_prediction}
\end{equation}
with $\xi_x^{\theta,s}(t_i)=x_i$. The clarity vector $q^{\theta,s}(t)\in(0,1]^{N_p}$ is propagated using~\eqref{eq:clarity_dynamics}, initialized at $q_i$. 
Thus, disturbances and filter interventions affect the predicted information acquisition.

\paragraph{Safety-aware filtering during planning.}
We now describe the approximate safety filter $\Fcal^{\mathrm{pred}}$ used in~\eqref{eq:controlled_prediction}, based on the robust policy control barrier function (RPCBF) approach~\citep{knoedler2025safety}. 
Let $h:\Xcal\rightarrow\R$ encode the admissible state set as $\Scal=\{x\in\Xcal:h(x)\leq0\}$,
 and let $\pi_b:\Xcal\rightarrow\Ucal$ denote a prescribed design policy satisfying the input constraints. At each predicted state $x=\xi_x^{\theta,s}(t)$ along a candidate trajectory, the RPCBF estimates the maximum future constraint violation by simulating $\pi_b$ under the sampled disturbance fields $d_i^{(s)}$. The sampled finite-horizon value function is
\begin{equation}
 \begin{gathered}
\widehat V_i(x)=\max_{s'=1,\ldots,N_d}
 \sup_{0\leq\tau\leq T_N}
 h\bigl(x_i^{\pi_b,s'}(\tau;x)\bigr),
 \end{gathered}
 \label{eq:rpcbf_value}
\end{equation}where $x_i^{\pi_b,s'}(\tau;x)\in\Xcal$ denotes the auxiliary state under $\pi_b$ and field $d_i^{(s')}$, initialized at $x_i^{\pi_b,s'}(0;x)=x$. The inner index $s'$ ranges over all $N_d$ fields while the outer rollout index $s$ is fixed. The sample set is held fixed when estimating the gradient with respect to $x$.
 Using $\widehat V_i(x)$ and its estimated gradient, the filter modifies the nominal tracking input $u_{\mathrm{nom}}^{\theta,s}(t)$ by minimizing its deviation subject to the RPCBF constraint and input bounds $u\in\Ucal$. 
For control-affine dynamics and polyhedral input bounds, imposing the sampled barrier conditions yields a quadratic program (QP)~\citep{knoedler2025safety}.
 The filtered input is used to propagate the predicted robot motion in~\eqref{eq:controlled_prediction}, with $\Fcal^{\mathrm{pred}}$ applied throughout each candidate rollout. 
The purpose of this filter is to account for how potential safety interventions affect the predicted motion and information acquisition, rather than to guarantee safety during planning.

\paragraph{Trajectory cost and optimization.}
Each candidate is evaluated using the accumulated clarity deficit along its predicted motion, with a soft penalty for state-constraint violations. Using the constraint function $h$ defined above and a penalty coefficient $\eta\in\Rnonneg$, the finite-horizon cost is
\begin{equation}
 \widehat J_i(\theta)=\frac1{N_d}\sum_{s=1}^{N_d}
 \int_{t_i}^{t_i+T_N}\left[
 q_d(q^{\theta,s}(t))+
 \eta[h(\xi_x^{\theta,s}(t))]_+^2
 \right]dt.
 \label{eq:sample_cost}
\end{equation}
The cost accounts for disturbance effects and safety-filter interventions through the predicted state and clarity trajectories. The geometric paths from Sec.~\ref{sec:geometric_init} initialize separate MBD optimizations using~\eqref{eq:mbd_forward}--\eqref{eq:mbd_reverse}, with $J=\widehat J_i$. Each optimization refines its candidate, and the lowest-cost evaluated candidate with a finite cost is selected, with parameters $\theta_i^\star\in\Theta$.

\paragraph{Execution and belief updates.}
The robot executes an initial segment of the selected candidate using the tracking controller and runtime filter $\Fcal^{\mathrm{exec}}$, which has the same argument and output spaces as $\Fcal^{\mathrm{pred}}$, then replans from the updated state and clarity.
 Actual disturbance measurements continue updating $\Bcal_i^d$ through POAM during both stages.

\begin{figure}[!htbp]
\centering
\includegraphics[width=\linewidth]{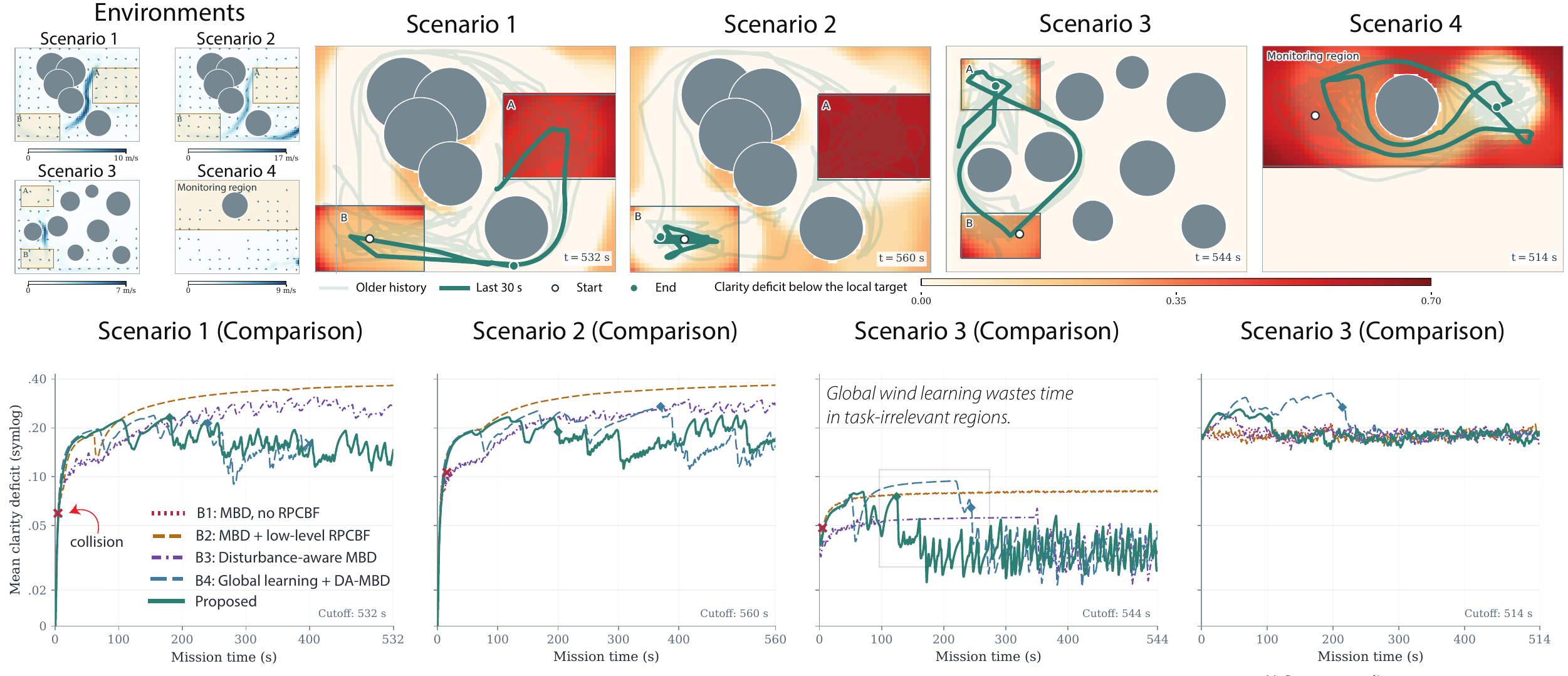}
\caption{
Monitoring across four scenarios. Top: environments and proposed trajectories; bottom: mean clarity-deficit histories for all methods. Comparison durations are $532$, $560$, $544$, and $514\,\mathrm{s}$ for Scenarios~1--4, respectively.}
\vspace{-5mm}
\label{fig:proposed_scenarios}
\end{figure}

\section{Results and Discussion}
\label{sec:results}
We evaluate four environments with spatially varying, time-invariant wind, using double-integrator nominal dynamics and a relative-air-velocity force model. Methods share each scenario's geometry, initial state and clarity, and MBD budget. The planning horizon is $15\,\mathrm{s}$, with $5\,\mathrm{s}$ execution intervals and no initial wind measurements. Appendix~\ref{app:experiments} provides simulation details.

\textbf{Baselines and evaluation:}
We compare \textbf{B1}, zero-wind MBD without filtering; \textbf{B2}, zero-wind MBD with the learned wind belief used in tracking and runtime RPCBF; \textbf{B3}, disturbance-aware MBD with RPCBF during planning and execution, learning only from measurements collected while monitoring; and \textbf{B4}, global wind learning followed by disturbance-aware MBD. The proposed method targets learning along monitoring-relevant paths before using the same planner as B4.

Figure~\ref{fig:proposed_scenarios} compares monitoring histories, while Table~\ref{tab:monitoring_results} in Appendix~\ref{app:experiments} reports the final mean clarity deficit $D(T)$ and its time average $D_T=T^{-1}\int_0^T D(t)\,dt$, 
with $D(t)=q_d(q(t))$ and elapsed time measured from the start of each run. Contact terminates a run, and monitoring deficits are not extrapolated beyond that point.

\textbf{Route-dependent disturbances:}
In Scenario~1, wind opposes travel from region B to A through the central passage, while a sheltered lower bypass is available. B2 and B3 do not enter A within the comparison window, whereas the proposed method does, reducing $D_T$ from B3's $0.224$ to $0.166$. B4 also reaches A but contacts an obstacle at $399.0\,\mathrm{s}$. 
In Scenario~2, wind also opposes travel through the lower passage, making the upper bypass important.
 Again, B2 and B3 do not enter A. Targeted learning ends at $200\,\mathrm{s}$ versus B4's $370\,\mathrm{s}$ and yields lower $D_T$, although B4 has the lower final deficit. The improvement therefore concerns accumulated monitoring performance, not every instantaneous deficit.

\textbf{Spatial allocation of learning:}
In Scenario~3, the proposed method learns disturbances along paths connecting the two monitoring regions rather than throughout the workspace. By focusing learning on monitoring-relevant paths, the proposed method reduces estimation time from $244$ to $124\,\mathrm{s}$ and lowers $D_T$ from $0.0540$ to $0.0418$ ($22.6\%$) relative to B4. In Scenario~4, the monitoring region is largely calm. Although the proposed method reduces learning time from $214$ to $102\,\mathrm{s}$ relative to B4, B2 achieves a lower $D_T$, showing that dedicated learning may be unnecessary when disturbances have little effect on monitoring.

\section{Conclusion}
\label{sec:conclusion}
We presented a two-stage framework combining monitoring-informed disturbance learning with disturbance-aware trajectory optimization. 
In the four reported runs, monitoring-informed sampling reduces dedicated learning time by $25.0$--$52.3\%$ relative to global estimation and improves monitoring when disturbances restrict access to sensing locations. The largely calm scenario also exposes the cost of unnecessary learning. 
\label{maintext:end}

\clearpage

\bibliography{l4dc2027-sample}

\appendix
\section{Algorithmic Details}
\label{app:algorithm}
This appendix details Algorithm~\ref{alg:framework}, which switches from disturbance estimation to persistent informative planning when the weighted disturbance uncertainty satisfies $U_i\leq\varepsilon_d$.

\subsection{Planning state and geometric initialization}
At planning time $t_i$, the available quantities are $x_i\in\Xcal$, $q_i\in(0,1]^{N_p}$, the disturbance data $\Dcal_i^d=\Dcal_{t_i}^d$, and their belief $\Bcal_i^d$. The geometry, target clarity $\bar q$, horizon $T_N$, uncertainty tolerance $\varepsilon_d$, and remaining design parameters are fixed inputs. The index $i\in\naturals$ counts planning iterations and is not reset between stages. 
The retained waypoint is $z\in\Pcal_{\mathrm{free}}$, or $z=\varnothing$ when none is available. The symbol $\bar\xi_i$ denotes the candidate selected for execution at $t_i$, or $\varnothing$ when no candidate is available.

The operation $\mathsf{GeoPaths}(x_i,q_i)$ predicts clarity deficits using~\eqref{eq:deficit_preview} and constructs geometrically distinct paths $\Gamma_i$, scored over their entire length by~\eqref{eq:path_score}. Numerical quadrature approximates the path integrals. These paths guide disturbance learning and initialize separate MBD optimizations; neither their construction nor scoring requires the disturbance belief.

The operation $\mathsf{PathWeights}(\Gamma_i,a_i)$ uses $\kappa_\rho$ in~\eqref{eq:learning_weights} to distribute path scores to nearby cells and normalize the weights $\omega_i$. It also returns their unnormalized sum $W_i$. If $W_i=0$, the algorithm sets $\omega_i=0$ and $U_i=+\infty$, preventing an undefined normalization or premature transition to monitoring.

\subsection{Disturbance measurement selection}
For positive total weight, the algorithm evaluates $U_i$ using~\eqref{eq:weighted_uncertainty}. If $U_i\leq\varepsilon_d$, it leaves the first loop before selecting another disturbance-measurement waypoint. Otherwise, $\mathsf{SelectWaypoint}$ retains a usable current waypoint or applies~\eqref{eq:measurement_selection} to the candidate set $\Zcal_i$. A new waypoint is needed after arrival, rejection of the connecting route, or lack of progress. Candidates are excluded when their routes intersect obstacles or are predicted to require inputs outside $\Ucal$. If $W_i=0$ or no usable candidate remains, the routine returns $\varnothing$.

The acquisition value $G_i(z)$ in~\eqref{eq:measurement_gain} predicts the reduction in weighted disturbance variance from one measurement, holding spatial weights and GP parameters fixed; subsequent POAM updates are not predicted. The travel penalty in~\eqref{eq:measurement_selection} accounts for distance, not the full downstream monitoring benefit. Measurements acquired during execution update the disturbance belief.

Given a waypoint, $\mathsf{MeasurementPath}(t_i,x_i,z,\Bcal_i^d)$ constructs a candidate trajectory $\bar\xi_i$ toward it using the known geometry and current disturbance belief. It returns $\varnothing$ if $z=\varnothing$ or no usable trajectory is available. A prescribed recovery policy is then used during execution. The specific recovery controller is an implementation choice, not an additional safety guarantee.

\subsection{Disturbance-aware MBD optimization}
After the transition, $\mathsf{Initialize}(\Gamma_i,x_i)$ time-parameterizes the geometric paths to form a finite set $\Theta_i^0\subset\Theta$ of candidate parameter vectors. Available hold and shifted previous candidates may also be retained. Each initialization is refined independently using the MBD equations in Sec.~\ref{sec:mbd}; candidate parameters from different chains are not averaged together.

The operation $\mathsf{MBD}(t_i,x_i,q_i,\Bcal_i^d,\Theta_i^0)$ evaluates each candidate using the sampled fields $d_i^{(s)}$, the filtered dynamics~\eqref{eq:controlled_prediction}, and the cost~\eqref{eq:sample_cost}. A sampled field is held fixed along its complete rollout. At each predicted state, the tracking input is modified by $\Fcal^{\mathrm{pred}}$, whose RPCBF value estimate is constructed under the design policy as in~\eqref{eq:rpcbf_value}. These design-policy predictions are auxiliary calculations for the filter; the candidate itself is propagated using the filtered tracking input. Clarity is propagated along the resulting state trajectory, and the disturbance belief remains fixed within the planning call.

The cost is averaged over disturbance samples before computing the MBD proposal weights. Initial candidates, sampled proposals, and refined means are evaluated before selection. If all proposal weights vanish, no weighted update is defined and that chain retains its previously evaluated candidates. Among candidates with defined, finite evaluations, the routine selects the lowest-cost trajectory and returns its parameters $\theta_i^\star\in\Theta$; if none is available, it returns $\varnothing$. A finite set of predicted trajectories does not certify physical safety. No hypothetical measurement is added to $\Dcal_i^d$ during this optimization.

\subsection{Execution and actual-data updates}
The operation $\mathsf{Execute}_i(\bar\xi_i)$ follows an initial segment of the candidate using the tracking controller and runtime filter $\Fcal^{\mathrm{exec}}$. If $\bar\xi_i=\varnothing$, the prescribed recovery controller supplies an input in $\Ucal$.
 In the first stage the candidate leads toward a disturbance-measurement waypoint; in the second it is selected by MBD. The routine returns the next planning time $t_{i+1}$, state $x_{i+1}$, clarity $q_{i+1}$, and new disturbance observations
\begin{equation*}
 \Delta\Dcal_i^d=\{(r_j,y_j^d):t_i<t_j^d\leq t_{i+1}\}.
\end{equation*}
The operation $\mathcal U_d$ incorporates the accumulated observations into the POAM model, including its online variational and kernel updates. The dataset union in Algorithm~\ref{alg:framework} records this assimilation and does not imply that the model must be retrained from scratch. Measurements of the quantities of interest also remain active, so clarity evolves throughout both stages. The planning clock, clarity, and disturbance belief are carried across the transition without resetting. If the uncertainty threshold is never met, the first stage continues; the algorithm assumes no terminal mission time.

\begin{algorithm2e}[p]
\small
\DontPrintSemicolon
\SetAlgoNoEnd
\KwIn{$t_0,x_0,q_0,\Bcal_0^d,\Dcal_0^d$; models, geometry, $\bar q$, and design parameters}
$i\gets0,\quad z\gets\varnothing$\;
\tcp{Stage 1: Disturbance estimation}
\While{\textnormal{true}}{
 $(\Gamma_i,a_i)\gets\mathsf{GeoPaths}(x_i,q_i)$\;
 $(\omega_i,W_i)\gets\mathsf{PathWeights}(\Gamma_i,a_i)$\;
 $U_i\gets+\infty$\;
 \lIf{$W_i>0$}{$U_i\gets\bigl(\sum_{p=1}^{N_p}\omega_{i,p}\operatorname{tr}P_d(r_p,t_i)\bigr)^{1/2}$}
 \lIf{$U_i\leq\varepsilon_d$}{\textbf{break}}
 $z\gets\mathsf{SelectWaypoint}(z,x_i,\omega_i,\Bcal_i^d)$\;
 $\bar\xi_i\gets\mathsf{MeasurementPath}(t_i,x_i,z,\Bcal_i^d)$\;
 $(t_{i+1},x_{i+1},q_{i+1},\Delta\Dcal_i^d)\gets\mathsf{Execute}_i(\bar\xi_i)$\;
 $\Dcal_{i+1}^d\gets\Dcal_i^d\cup\Delta\Dcal_i^d$\;
 $\Bcal_{i+1}^d\gets\mathcal U_d(\Dcal_{i+1}^d)$\;
 $i\gets i+1$\;
}
\BlankLine
\tcp{Stage 2: Disturbance-aware informative planning}
\While{\textnormal{true}}{
 $(\Gamma_i,a_i)\gets\mathsf{GeoPaths}(x_i,q_i)$\;
 $\Theta_i^0\gets\mathsf{Initialize}(\Gamma_i,x_i)$\;
 $\theta_i^\star\gets\mathsf{MBD}(t_i,x_i,q_i,\Bcal_i^d,\Theta_i^0)$\;
 $\bar\xi_i\gets\varnothing$\;
 \lIf{$\theta_i^\star\neq\varnothing$}{$\bar\xi_i\gets\bar\xi(\cdot;\theta_i^\star)$}
 $(t_{i+1},x_{i+1},q_{i+1},\Delta\Dcal_i^d)\gets\mathsf{Execute}_i(\bar\xi_i)$\;
 $\Dcal_{i+1}^d\gets\Dcal_i^d\cup\Delta\Dcal_i^d$\;
 $\Bcal_{i+1}^d\gets\mathcal U_d(\Dcal_{i+1}^d)$\;
 $i\gets i+1$\;
}
\caption{Persistent disturbance-aware informative planning. The estimation stage ends only when weighted posterior uncertainty meets its tolerance.}
\label{alg:framework}
\end{algorithm2e}

\clearpage
\section{Environments, Baselines, and Detailed Results}
\label{app:experiments}
\begin{figure}[!ht]
\centering
\includegraphics[width=\linewidth,height=0.78\textheight,keepaspectratio]{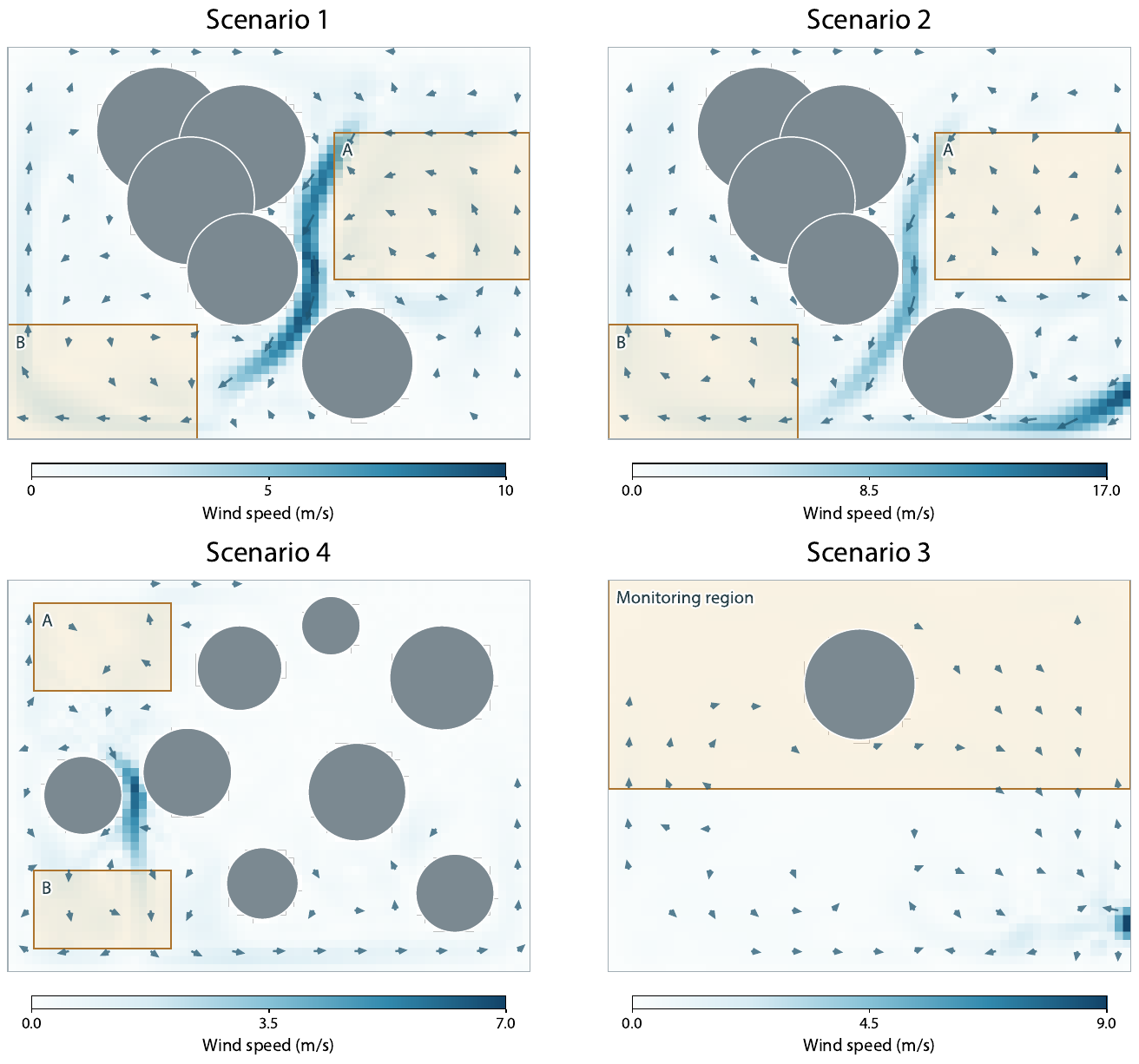}
\caption{The four evaluation environments. 
Scenarios~1 and~2 test alternative passages under opposing wind; Scenario~3 has two monitoring regions within a larger workspace; Scenario~4 has a largely calm monitoring region. The lower-left and lower-right panels correspond to Scenarios~3 and~4 in the main text; their embedded labels are reversed.}
\label{fig:environments-two-by-two}
\end{figure}

\begin{figure}[!ht]
\centering
\includegraphics[width=\linewidth,height=0.78\textheight,keepaspectratio]{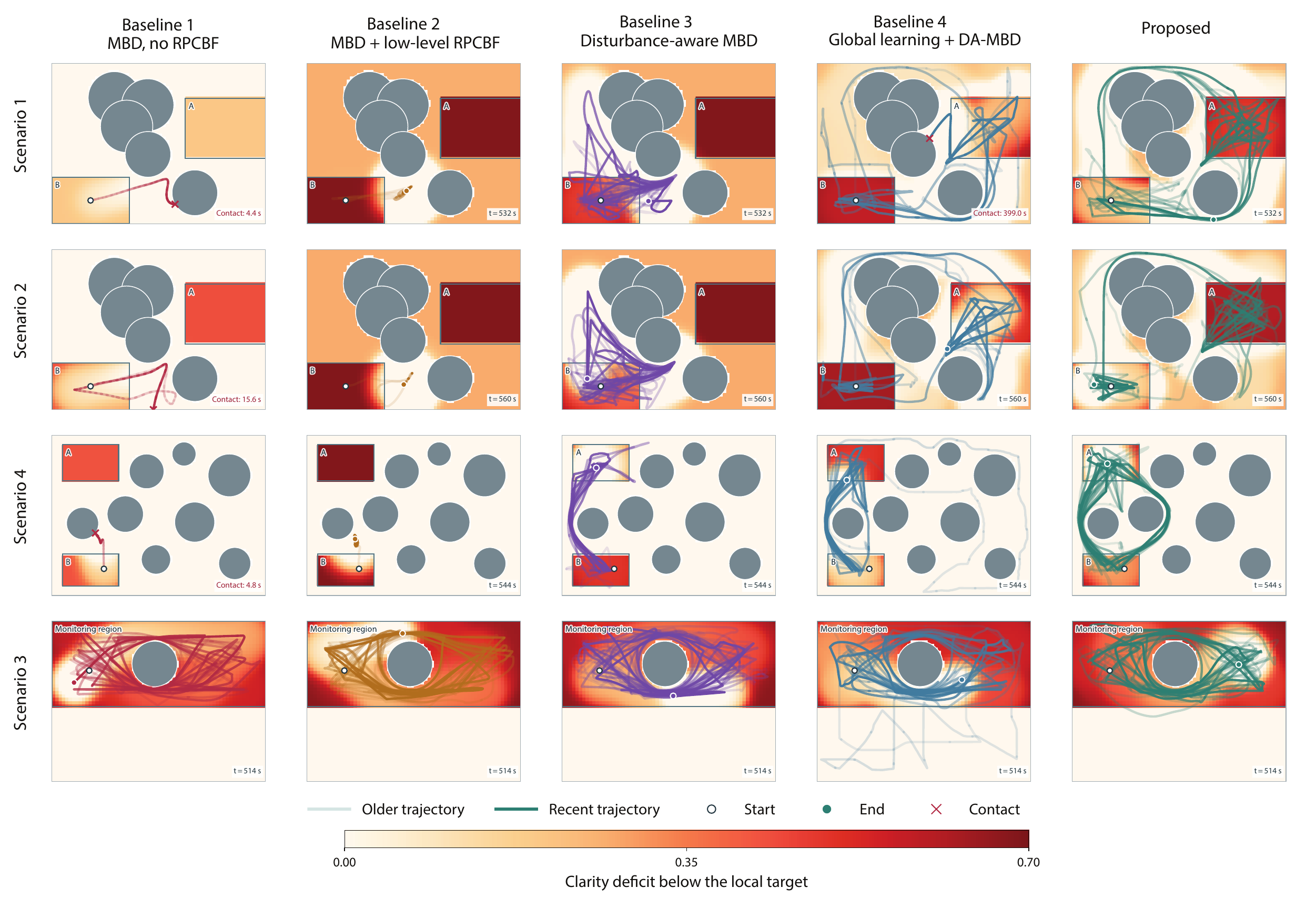}
\caption{
Trajectories of the proposed method and baselines, with rows corresponding to Scenarios~1--4 in the main text. The embedded labels in the final two rows are reversed. Runs ending in contact are terminated, not completed comparisons; Table~\ref{tab:monitoring_results} reports the numerical results.}
\label{fig:all-method-trajectories}
\end{figure}

Four scenarios assess disturbance-aware trajectory prediction, alternative-route discovery, and monitoring-informed disturbance learning. Wind affects robot motion while the QoI dynamics remain separate. Figures~\ref{fig:environments-two-by-two}, \ref{fig:proposed_scenarios}, and \ref{fig:all-method-trajectories} show the environments and trajectory comparisons. Simulations were performed on an Apple M3 MacBook Air with $16\,\mathrm{GB}$ of unified memory.

\subsection{Simulation and comparison protocol}
\label{app:protocol}
Simulations use double-integrator nominal dynamics with wind entering through a relative-air-velocity force model. Wind varies spatially but remains fixed in time. Methods share each scenario's geometry, initial state and clarity, and MBD budget. The planning horizon is $15\,\mathrm{s}$, with $5\,\mathrm{s}$ execution intervals. Wind estimation begins without measurements, and POAM is updated online from observations collected during execution.

Dedicated learning terminates when the path-weighted disturbance uncertainty in~\eqref{eq:weighted_uncertainty} falls below $\varepsilon_d=1.10\,\mathrm{m/s}$. B4 instead evaluates disturbance uncertainty throughout the free workspace. This posterior uncertainty criterion is model-dependent and does not bound the true wind-estimation error.

Scenarios~1--4 are evaluated over $532$, $560$, $544$, and $514\,\mathrm{s}$, respectively, measured from mission start and including dedicated learning. QoI sensing remains active during learning. These evaluation windows do not impose a terminal time on the persistent planner. Runs end upon contact, and monitoring deficits are not extrapolated thereafter.

\subsection{Comparison methods}
\label{app:baselines}
\paragraph{B1: MBD without RPCBF.}
B1 assumes zero wind in planning and executes MBD trajectories without safety filtering.

\paragraph{B2: MBD with runtime RPCBF.}
B2 uses zero-wind planning, while tracking and runtime RPCBF use the learned wind belief. The planner therefore does not account for disturbance-induced motion or filter corrections when choosing trajectories.

\paragraph{B3: Disturbance-aware MBD without dedicated learning.}
B3 begins informative monitoring immediately, evaluating candidate trajectories using the learned wind belief and RPCBF during planning and execution. POAM learns from measurements collected during monitoring, without a dedicated disturbance-learning stage.

\paragraph{B4: Global learning followed by disturbance-aware MBD.}
B4 reduces wind uncertainty throughout the free workspace, including regions irrelevant to monitoring, before using the same disturbance-aware MBD planner as the proposed method.

\paragraph{Proposed: Monitoring-informed learning followed by disturbance-aware MBD.}
The proposed method prioritizes disturbance measurements near monitoring-relevant geometric paths, including transit passages, before using the same disturbance-aware MBD planner as B4. Learning allocation is handled in the first stage rather than by adding a disturbance-information reward to the monitoring cost.

\subsection{Metrics and learning durations}
For each simulation, define the mean clarity deficit and its time average as
\begin{equation}
 D(t)=\frac{1}{N_p}\sum_{p=1}^{N_p}[\bar q_p-q_p(t)]_+,
 \qquad D_T=\frac{1}{T}\int_0^T D(t)\,dt.
 \label{eq:experimental_metrics}
\end{equation}
The final deficit $D(T)$ depends on the revisit cycle, whereas $D_T$ captures monitoring loss over the full comparison window. Table~\ref{tab:monitoring_results} reports both; bold denotes the lowest $D_T$ among completed runs, not a statistically significant difference.

\begin{table}[htbp]
\centering
\small
\caption{
Monitoring performance: $D_T$, with final deficit $D(T)$ in parentheses. Bold indicates the lowest $D_T$ among completed runs; contact times are reported for terminated runs.}
\label{tab:monitoring_results}
\begin{tabular}{lcccc}
\toprule
Method & Scenario 1 & Scenario 2 & Scenario 3 & Scenario 4\\
 & $T=532\,\mathrm{s}$ & $T=560\,\mathrm{s}$ & $T=544\,\mathrm{s}$ & $T=514\,\mathrm{s}$\\
\midrule
B1 & Contact: 4.4 s & Contact: 15.6 s & Contact: 4.8 s & 0.181 (0.161)\\
B2 & 0.293 (0.366) & 0.301 (0.367) & 0.078 (0.081) & \textbf{0.180} (0.180)\\
B3 & 0.224 (0.280) & 0.221 (0.267) & 0.047 (0.035) & 0.186 (0.184)\\
B4 & Contact: 399.0 s & 0.190 (0.156) & 0.054 (0.040) & 0.221 (0.177)\\
Proposed & \textbf{0.166} (0.146) & \textbf{0.177} (0.171) & \textbf{0.042} (0.027) & 0.189 (0.189)\\
\bottomrule
\end{tabular}

\end{table}

Table~\ref{tab:learning_times} reports simulated disturbance-learning times and percentage reductions $100(t_{\mathrm{B4}}-t_{\mathrm{prop}})/t_{\mathrm{B4}}$, where $t_{\mathrm{B4}}$ and $t_{\mathrm{prop}}$ are the respective durations. These are mission times, not computation times; QoI sensing remains active throughout learning.
\begin{table}[htbp]
\centering
\caption{Dedicated wind-estimation time.}
\label{tab:learning_times}
\begin{tabular}{lrrr}
\toprule
Scenario & B4 (s) & Proposed (s) & Reduction\\
\midrule
1 & 240 & 180 & $25.0\%$\\
2 & 370 & 200 & $45.9\%$\\
3 & 244 & 124 & $49.2\%$\\
4 & 214 & 102 & $52.3\%$\\
\bottomrule
\end{tabular}

\end{table}

\subsection{Scenario 1: Central opposing wind and a lower bypass}
Wind opposes travel from region B to A through the central passage, while a sheltered lower bypass remains available. B2 and B3 do not reach A within $532\,\mathrm{s}$, although B3 achieves a lower mean deficit than B2. The proposed method reaches A and reduces $D_T$ to $0.166$ versus $0.224$ for B3. B4 also reaches A but contacts an obstacle at $399.0\,\mathrm{s}$; it therefore has no full-window monitoring score.

\subsection{Scenario 2: Opposing wind in the central and lower passages}
Wind also opposes the lower passage, making the upper bypass important. B2 and B3 again fail to reach A within the $560\,\mathrm{s}$ comparison window. Both dedicated-learning methods reach A. Targeted learning takes $200\,\mathrm{s}$ versus $370\,\mathrm{s}$ for B4. The proposed method reduces $D_T$ from $0.190$ to $0.177$, although B4 has the lower final deficit ($0.156$ versus $0.171$).

\subsection{
Scenario 3: Monitoring concentrated in two regions}
Two monitoring regions are embedded in an obstacle-filled workspace with additional wind outside monitoring-relevant routes. B2 never reaches A, while B3 reaches it only at $351.55\,\mathrm{s}$, illustrating the delay in discovering an alternative without dedicated disturbance learning. Compared with B4, targeted learning ends after $124$ rather than $244\,\mathrm{s}$, leaving more time for monitoring. The proposed method reduces $D_T$ from $0.0540$ to $0.0418$ ($22.6\%$) and $D(T)$ from $0.0396$ to $0.0269$ (about $32\%$). B4 reaches A earlier during learning ($54.10$ versus $74.65\,\mathrm{s}$), so the proposed method's improvement reflects a shorter learning stage rather than an earlier first arrival.

\subsection{
Scenario 4: Calm monitoring region and remote wind}
The monitoring region is largely calm, with stronger wind elsewhere. Targeted learning takes $102\,\mathrm{s}$ versus B4's $214\,\mathrm{s}$ and lowers $D_T$ from $0.221$ to $0.189$, although its final deficit is higher ($0.189$ versus $0.177$). However, B2 attains the lowest $D_T=0.180$ (B1: $0.181$; B3: $0.186$). Dedicated learning may therefore be unnecessary when disturbances have little effect on monitoring, and the fixed switching threshold still incurs learning cost in this scenario.

\subsection{Planning time and interpretation}
In Scenario~3, median planning times after compilation are $1.64\,\mathrm{s}$ (proposed) and $1.60\,\mathrm{s}$ (B4), with 95th percentiles of $1.93$ and $1.73\,\mathrm{s}$, respectively. Both methods use the same MBD budget. Timings include posterior and field preparation, geometric initialization, optimization, and safety-belief refresh. The baselines distinguish execution-time compensation (B2), disturbance-aware monitoring without dedicated learning (B3), and spatial allocation of learning with a common monitoring planner (B4 versus proposed). These comparisons do not separately isolate geometric initialization, weighting, or the switching rule. Each scenario reports one run. The proposed method completes all four comparison windows without contact, but infeasible RPCBF steps occur, and B4 contacts an obstacle in Scenario~1 despite filtering. The results do not establish guaranteed collision avoidance or statistical robustness. The path weights approximate task relevance rather than the exact long-term value of disturbance information.

\clearpage
\clearpage

\end{document}